\documentclass{article}
\usepackage[utf8]{inputenc}
\usepackage{graphicx}
\usepackage[round,authoryear]{natbib}
\usepackage[a4paper, margin=1in]{geometry}
\usepackage[utf8]{inputenc}
\usepackage[T1]{fontenc}
\usepackage{microtype}
\usepackage[hyphens]{url}
\usepackage{color}
\definecolor{darkblue}{rgb}{0.0, 0.0, 0.55}
\usepackage[colorlinks=true,linkcolor=darkblue,citecolor=darkblue,urlcolor=darkblue]{hyperref} 
\usepackage{comment}
\usepackage{booktabs}
\usepackage{longtable}
\usepackage{pdflscape}
\usepackage{geometry}
\usepackage{amsmath}
\usepackage{listings}
\usepackage{array}
\usepackage{caption}
\usepackage{xspace}
\usepackage[capitalize,nameinlink,noabbrev]{cleveref}[0.21]
\usepackage{inconsolata}

\definecolor{quotebg}{RGB}{245,245,250}
\definecolor{quoteborder}{RGB}{110,120,180}
\lstdefinestyle{feedbackprompt}{
  basicstyle=\footnotesize\ttfamily,
  backgroundcolor=\color{quotebg},
  frame=single,
  framesep=8pt,
  framerule=1pt,
  rulecolor=\color{quoteborder},
  breaklines=true,
  breakatwhitespace=false,
  captionpos=b,
  showstringspaces=false,
  numbers=none,
  xleftmargin=5pt,
  xrightmargin=5pt,
  aboveskip=10pt,
  belowskip=10pt
}

\usepackage{libertine}
\newcommand{\glove}{GloVe\xspace}
\newcommand{\sdat}{S-DAT\xspace}
\newcommand{\dat}{DAT\xspace}
\newcommand{\textthreelarge}{text-embedding-3-large\xspace}
\newcommand{\nomic}{nomic-embed-text-v1.5\xspace}
\newcommand{\granite}{granite-embedding-278m-multilingual\xspace}
\newcommand{\snowflake}{snowflake-arctic-embed-l\xspace}
\newcommand{\snowflaketwo}{snowflake-arctic-embed-l-v2.0\xspace}
\newcommand{\xlmroberta}{XLM-RoBERTa\xspace}

\usepackage{pifont}
\newcommand{\checkmark}{\textcolor{green}{\ding{51}}}
\newcommand{\xmark}{\textcolor{red}{\ding{55}}}

\title{S-DAT: A Multilingual, GenAI-Driven Framework for Automated Divergent Thinking Assessment}

\author{
    Jennifer Haase\\
    Weizenbaum Institute and Humboldt University\\ Berlin, Germany\\
    \url{jennifer.haase@hu-berlin.de}
    \and
    Paul H. P. Hanel\\
    University of Essex, Colchester, UK\\
    \url{p.hanel@essex.ac.uk}
    \and
    Sebastian Pokutta\\
    TU Berlin and Zuse Institute Berlin \\ Berlin, Germany\\
    \url{pokutta@zib.de}
}

\date{May 14, 2025}

\begin{document}

\maketitle

\begin{abstract}
This paper introduces \sdat (Synthetic-Divergent Association Task), a scalable, multilingual framework for automated assessment of divergent thinking (DT)---a core component of human creativity. Traditional creativity assessments are often labor-intensive, language-specific, and reliant on subjective human ratings, limiting their scalability and cross-cultural applicability. In contrast, \sdat leverages large language models and advanced multilingual embeddings to compute semantic distance---a language-agnostic proxy for DT. We evaluate \sdat across eleven diverse languages, including English, Spanish, German, Russian, Hindi, and Japanese (Kanji, Hiragana, Katakana), demonstrating robust and consistent scoring across linguistic contexts. Unlike prior DAT approaches, the \sdat shows convergent validity with other DT measures and correct discriminant validity with convergent thinking. This cross-linguistic flexibility allows for more inclusive, global-scale creativity research, addressing key limitations of earlier approaches. \sdat provides a powerful tool for fairer, more comprehensive evaluation of cognitive flexibility in diverse populations and can be freely assessed online: \url{https://sdat.iol.zib.de/}.

\end{abstract}

\section{Introduction}

Large language models (LLMs) have become powerful tools in creativity research, increasingly used to support human ideation across domains such as mathematical problem-solving \citep{ye_assessing_2025}, scientific discovery \citep{gottweisAcceleratingScientificBreakthroughs2025, boikoEmergentAutonomousScientific2023}, and music composition \citep{bodily_operationalizing_2024}. Within the growing field of \textit{computational creativity}, AI systems have also demonstrated the ability to produce content considered creative, ranging from visual art \citep{dipaola_using_2016} to product design \citep{kaila_gardening_2024} and architecture \citep{roncoroni_computational_2024}.

Beyond the generation of creative artifacts, AI is now being harnessed to assess creativity itself. This marks a fundamental shift in how we conceptualize and measure creative cognition: from human-judged, often subjective methods, toward scalable, automated assessments using computational proxies. For instance, recent tools compute semantic distance between ideas to approximate the cognitive flexibility required in divergent thinking (DT, the cognitive process of generating multiple, varied, and original ideas in response to open-ended problems)---a core component of creativity \citep{olson_naming_2021, organisciak_beyond_2023, cropley_automated_2025}. In parallel, models such as reinforcement learning agents have been shown to autonomously explore symbolic action spaces, suggesting that creativity can be formalized as structured conceptual exploration \citep{jinCreativityAIAutomatic2022}.

Yet, despite these advances, the measurement of creativity---particularly DT---remains constrained by longstanding challenges. Common tasks like the \textit{Alternate Uses Test} (AUT, \citealt{christensen_alternate_1960}) or story generation exercises are cognitively demanding and time-intensive for participants \citep{mohamed_creative_2011, chaudhuri_evaluating_2025}, and typically require labor-intensive human scoring via frameworks like the \textit{Consensual Assessment Technique} (CAT, \citealt{amabile_assessing_1996}). Moreover, most established tools are designed in English and rely on language-specific embeddings such as GloVe \citep{olson_naming_2021}, limiting their accessibility for non-English-speaking populations and undermining their validity in cross-cultural research.

Recent advances in generative AI (GenAI), particularly in multilingual and context-aware embeddings, offer promising alternatives. Approaches leveraging semantic distance---defined as the dissimilarity between word meanings in high-dimensional vector space---are increasingly used to approximate DT across languages \citep{kenett_what_2019, hass_tracking_2017, heinen_semantic_2018}. Notably, multilingual adaptations of creativity tasks have begun to emerge \citep{patterson_multilingual_2023, luchini_automated_2025}, yet they rely on models not optimized for fine-grained semantic comparison. For instance, models such as \xlmroberta prioritize general language understanding, which may introduce noise in creativity assessments, especially across diverse scripts and linguistic structures. Furthermore, high-burden tasks like AUT or creative story writing restrict the number of trials and may introduce confounds such as task fatigue or writing ability. Simpler, low-effort tasks—such as the \textit{Divergent Association Task} (DAT), which prompts participants to generate semantically unrelated words—provide a more efficient and accessible format for assessing DT \citep{olson_naming_2021}. However, the original DAT relies on English-only static embeddings and does not support multilingual administration or cross-linguistic comparability.

In this paper, we propose \sdat, a multilingual, GenAI-powered framework for the automated assessment of DT. Our approach builds on the DAT format, but leverages transformer-based embedding models to compute semantic distance across a wide range of scripts and languages. Rather than relying on translation or language-specific calibration, \sdat operates natively in multiple languages by drawing on Unicode-based encodings and broad tokenization.

From both a scientific and ethical perspective, \sdat addresses the need for fairer, more inclusive tools in creativity research. By eliminating reliance on English and minimizing participant burden, it enables large-scale, cross-linguistic studies without privileging specific cultural or linguistic groups. In doing so, \sdat contributes to a broader shift toward equitable, globally relevant AI research—aligned with current debates on language bias and representational fairness in AI systems \citep{houNaturalLanguageProcessing2025, gallegosBiasFairnessLarge2024}.

To develop and validate \sdat, we benchmarked eight multilingual embedding models, identifying IBM's \textit{\granite} model as the most robust for semantic distance scoring across eleven languages and scripts. We calibrated the resulting scores against the original DAT \citep{olson_naming_2021} and validated \sdat using human-rated tasks such as the AUT and Bridge-the-Associative-Gap Task. Our results demonstrate that \sdat offers consistent and reliable scoring across diverse linguistic settings, providing a scalable and ethically grounded tool for DT assessment.

\section{Background and Motivation}
\subsection{Divergent Thinking as an Essential Aspect of Creativity}

Creativity, defined as the generation of novel and useful ideas \citep{runco_standard_2012, plucker_generalization_2004}, relies heavily on DT \citep{guilford_creativity_1950, runco_divergent_2010}, which is typically assessed using tasks like the AUT \citep{christensen_alternate_1960} or DAT \citep{olson_naming_2021}.

DT is a foundational indicator of creative potential. Meta-analyses confirm a modest link to creativity, with correlations around \textit{r}=.27 \citep{da_costa_personal_2015} and shared variance as low as 3\% \citep{said_metwaly_divergent_2024}. These rather moderate statistics also signal a critical limitation: DT is not synonymous with creativity. It captures one aspect---ideational fluency and variety---but does not encompass other essential components such as idea evaluation, contextual relevance, or implementation. Indeed, creativity often emerges from a dynamic interplay between divergent and convergent thinking. While DT expands the space of possibilities, convergent thinking involves narrowing those possibilities down to a viable, coherent solution using logic and critical reasoning \citep{cropley_praise_2006, runco_divergent_2012}. Guilford’s foundational research captured this contrast, coining the terms ``divergent production'' and ``convergent production'' to describe how people either explore new channels of thought or draw upon conventional solutions when solving problems \citep{guilford_creativity_1967}. Both modes are essential: creativity requires not just the generation of ideas, but also the ability to refine, select, and develop them.

Against this backdrop, we position our tool development not as a comprehensive measure of creativity, but as a focused proxy for one essential aspect: the ability to think in semantically diverse and varied ways. Our goal is to contribute to the broader creativity debate by offering a scalable, multilingual assessment of DT---one pillar among many in the architecture of creative cognition. Thus, while our approach centers on DT and semantic distance as a quantifiable entry point, it is situated within a broader recognition that creativity involves an interplay of divergent, convergent, emergent, and improvisational modes of thinking \citep{cromwell_creative_2023}. We offer the \sdat as a tool to probe this foundational capacity, not to define creativity in its entirety, but to enrich how one dimension of it can be understood and assessed across linguistic and cultural contexts.

\subsection{From Human Scoring to Automated Assessment of Creativity}

Creativity assessment has historically relied on labor-intensive and language-dependent methodologies. Traditional tools such as the AUT and the \textit{Torrance Tests of Creative Thinking} have shaped our understanding of creative cognition, particularly through their focus on DT. These tasks are typically scored along dimensions like fluency, originality, flexibility, and elaboration---most often by human raters, using either subjective criteria or the CAT \citep{silvia_assessing_2008, amabile_assessing_1996}. Despite strong validity, subjective human scoring is costly, slow, and prone to inter-rater inconsistency, limiting large-scale or cross-linguistic research \citep{patterson_multilingual_2023}.

To overcome these limitations, creativity researchers have increasingly turned to automated methods, particularly those powered by LLMs. A key innovation in this domain is the use of semantic distance---a computational measure of how conceptually distinct two ideas are within a high-dimensional embedding space. Tools like SemDis \citep{beaty_automating_2021} and the \dat \citep{olson_naming_2021} automate this process by analyzing the average semantic distance between generated words, offering a scalable proxy for ideational breadth. However, until recently, these tools were largely limited to English. A study by Patterson et al. \citeyearpar{patterson_multilingual_2023} represents a breakthrough, demonstrating that multilingual semantic distance---calculated using models such as XLM-RoBERTa and Multilingual BERT (mBERT)---can correlate with human creativity ratings across 12 languages. Their validation included correlations with creative traits like openness and self-reported achievements, supporting cross-cultural generalizability and opening new avenues for non-English creativity assessment.

Simultaneously, efforts to enhance scoring precision have evolved from embedding-based methods to fine-tuned LLMs. Organisciak et al. \citeyearpar{organisciak_beyond_2023} show that models trained directly on human-rated AUT responses outperform standard expert ratings, achieving correlations up to \textit{r}~=~.81 with human judgments. These models generalize well across tasks and even outperform zero-shot GPT-4 evaluations, challenging the ceiling of current AI-based assessments. Parallel work by Luchini et al. \citeyearpar{luchini_automated_2025} expanded automated creativity assessment to narrative tasks, showing that LLMs trained on multilingual stories could reliably predict human originality scores across 11 languages. Notably, even a monolingual English-trained model performed strongly on translated texts, highlighting the potential of LLMs for robust multilingual scoring. Nevertheless, these tools are not without vulnerabilities. Doebler et al. \citeyearpar{doebler_assessing_2025} emphasize the need for robustness checks in automated systems by introducing adversarial examples---responses with subtle semantic changes that drastically shift model scores. Their findings highlight model sensitivity and recommend adversarial training as a strategy to bolster score reliability.

Beyond text-based measures, LLMs have also been applied to figural and domain-specific tasks. Cropley and Marrone \citeyearpar{cropley_automated_2025} employed deep learning for automated scoring of drawing-based tests like the \textit{Test for Creative Thinking - Drawing Production} (TCT-DP, \citealt{urbanAssessingCreativityTest2005}), demonstrating feasibility and accuracy. Likewise, Goecke et al. \citeyearpar{goecke_automated_2024} showed that XLM-RoBERTa was strongly associated with originality ratings in a German-language scientific ideation task, \textit{r}~=~.80. Moreover, researchers like Kern et al. \citeyearpar{kern_assessing_2023} are refining prompt engineering strategies for GPT-4 to assess novelty, feasibility, and value in AUT responses in Japanese. Their work emphasizes the importance of task design and prompt structure in eliciting and evaluating creative responses across cultural contexts.

\subsection{Semantic Distance as a Proxy for Divergent Thinking}

Semantic distance provides a computational approximation of conceptual divergence. If a person generates ideas that are semantically far apart---such as ``giraffe'' and ``quantum mechanics''---this is interpreted as evidence of cognitive flexibility, a key component of DT. Semantic distance approaches in creativity research are based on the principle that highly creative ideas involve linking semantically distant concepts. The greater the dissimilarity between generated ideas, the higher the inferred cognitive flexibility and originality of the individual \citep{kenett_what_2019, hass_tracking_2017, heinen_semantic_2018}. 
DT tasks, such as the DAT, leverage this by instructing participants to generate words or concepts that are as unrelated as possible. Semantic distance thus operationalizes creativity by assessing how ``far apart'' in meaning the generated items are within a high-dimensional semantic space \citep{olson_naming_2021, hass_tracking_2017}.

Semantic distance is computed by mapping words to vectors in a high-dimensional space, using models such as \textit{Latent Semantic Analysis} (LSA, \citealt{hass_tracking_2017}), Word2Vec, and GloVe \citep{doebler_assessing_2025}. Approaches to semantic distance measurement include:
\begin{itemize}
    \item \textbf{Static Embedding Models} (e.g., LSA, Word2Vec, GloVe): Used fixed vector spaces based on co-occurrence statistics but suffered from context insensitivity and elaboration biases \citep{olson_naming_2021, doebler_assessing_2025}.
    \item \textbf{Dynamic Embedding Models} (e.g., BERT, RoBERTa): Produce context-sensitive representations, better capturing nuanced meaning in different languages and scripts \citep{patterson_multilingual_2023}.
    \item \textbf{Single-Item vs. Aggregate Scoring}: Studies differ in whether they compute pairwise distances between all generated words or select the maximum distance to represent semantic spread \citep{hass_tracking_2017, luchini_automated_2025}.
    \item \textbf{Adversarial Robustness Testing}: Automated scoring models can be sensitive to strategic manipulations, such as synonym substitution, necessitating robustness checks through adversarial examples \citep{doebler_assessing_2025}.
\end{itemize}
Thus, measurement approaches have evolved from simple unsupervised word comparisons to sophisticated models better aligned with human creativity ratings.

LLMs encode semantic relationships through training on vast text corpora, developing context-aware vector spaces where related words cluster together and unrelated words are distant. Recent research indicates that transformer-based LLMs develop universal feature spaces across languages and domains, making them highly suited for measuring semantic distance in multilingual and multicultural creativity assessments \citep{lan_sparse_2025, luchini_automated_2025}. Sparse autoencoders reveal that LLMs encode similar semantic structures despite varied training data, supporting cross-linguistic generalization \citep{lan_sparse_2025}.

LSA and GloVe embeddings are static, assuming fixed word meanings across contexts, leading to issues such as elaboration bias, where longer responses artificially inflate distance scores \citep{doebler_assessing_2025}. Modern transformer-based embeddings like mBERT and XLM-R solve these issues by producing dynamic, context-sensitive vector representations. These models not only allow for more robust assessments within a language but also across multiple languages without requiring translation \citep{patterson_multilingual_2023, luchini_automated_2025}. In multilingual settings, two strategies have been applied: translating all responses into English and scoring with English-based models \citep{luchini_automated_2025}, or directly employing multilingual embedding models like mBERT and XLM-R \citep{patterson_multilingual_2023}.
While semantic distance provides an efficient and scalable proxy for DT, it is important to acknowledge its limitations. Adversarial manipulation of inputs, elaboration bias, and differences in cultural semantic norms can all affect measurement validity \citep{doebler_assessing_2025}. These considerations become even more critical in languages with compound constructions (e.g., German) or logographic scripts (e.g., Japanese), where tokenization quality can significantly impact embedding accuracy and thus semantic distance measures. Careful calibration, multilingual benchmarking, and robustness testing are therefore essential for reliable application across diverse contexts.

\subsection{Prior DAT Scoring Approach} 

\citet{olson_naming_2021} built an accessible free-online tool to apply the DAT, which uses a \glove model to compute the semantic distance between all pairs of words in the generated word set (10 words), and then computes the average semantic distance as the final score by averaging the dissimilarity scores for the top seven words. The distance between two words is calculated as the cosine distance (or similarity) between their vector representations, where a smaller cosine similarity (or larger cosine distance) indicates greater semantic dissimilarity \citep{patterson_multilingual_2023}; see section on semantic distances for details.

The scoring strategy in \dat and related tasks involves computing the average pairwise semantic distance across all possible word pairs in a participant's set (e.g., ten words yield 45 pairs, \citealt{olson_naming_2021}). For responses containing multiple words, which is especially relevant for languages frequently using compound words like German (e.g., \textit{Datenschutzgrundverordnung}, meaning \textit{General Data Protection Regulation}), subword tokenization (e.g., Byte-Pair Encoding or WordPiece) may split compounds into smaller units, but the resulting embedding may not fully capture the holistic meaning. This can affect the accuracy of semantic distance computations unless the model was trained on sufficient compound-rich data and this is precisely where newer context-sensitive embedding models have advantages over static embedding models as an embedding is not simply ``the sum of the embeddings of its tokens'' but rather the embeddings on the specific token configuration, see, e.g., \citet{awasthyGraniteEmbeddingModels2025}. We refer the interested reader to \citet{schmidtTokenizationMoreCompression2024, doebler_assessing_2025, patterson_multilingual_2023} for more details and in-depth discussion, as tokenization is a topic in its own right beyond the scope of this paper.

While the original scoring approach for the \dat represented a major advance in automating creativity assessment, it is important to recognize several key limitations inherent to this method:

\begin{itemize}
    \item \textbf{Global, context-free embeddings:} The \glove model produces static, global word embeddings that are learned from co-occurrence statistics across a large corpus (Common Crawl). Each word is represented by a single vector, regardless of its context or sense. This means that polysemous words (e.g., ``bank'' as a financial institution vs. river bank) are mapped to the same point in the embedding space, potentially obscuring nuanced semantic relationships relevant to creativity.
    \item \textbf{Lack of contextualization:} Because \glove embeddings are not context-sensitive, they cannot capture the way meaning shifts depending on usage or surrounding words and also those of compound words. This is a known limitation for tasks where context and subtle distinctions are important, and it may reduce the sensitivity of the \dat to certain forms of creative association; this might be of particular relevance when composite words are input, e.g., ``ice cream'' and ``bus stop.''
    \item \textbf{English-only limitation:} The \glove model used in the original \dat is trained exclusively on English-language data. As a result, the scoring approach is not directly applicable to responses in other languages, severely limiting the potential for cross-linguistic or multicultural creativity assessment. This is particularly problematic for global research or for participants whose primary language is not English.
    \item \textbf{No cross-lingual comparability:} Since the \glove embedding is monolingual (i.e., English only), there is no principled way to compare or aggregate \dat scores across different languages. This precludes meaningful cross-cultural studies and makes it difficult to generalize findings beyond English-speaking populations.
\end{itemize}

These limitations motivate the need for more advanced, multilingual, and context-aware embedding models in the next generation of DAT scoring systems. By leveraging transformer-based models that provide contextualized and language-agnostic embeddings, it becomes possible to address these shortcomings and enable robust, scalable, and fair creativity assessment across diverse linguistic and cultural backgrounds. We therefore introduce \sdat, a multilingual, GenAI-powered scoring system that leverages contextual embeddings to overcome static, language-bound approaches. In the following section, we describe its development, calibration, and validation.

\section{\sdat development}
\label{sec:sdat}

The main purpose of the development of the \sdat is to overcome the aformentioned limitations and design a modern variant that satisfies the following key design criteria: (a) it is multilingual with (potential) support for a wide variety of scripts and languages, (b) it goes beyond static embeddings with static world lists allowing for broader inputs, (c) it is compatible with the original \dat, both in scores and score distributions, as well as being highly correlated on identical inputs, and (d) we want to avoid intra- and inter-language calibrations and it is desirable to avoid recalibration for newly added languages later down the road. To achieve these goals, we analyze several state-of-the-art multilingual transformer-based embedding models, calibrate them against the \dat (for score compatibility), test the scoring stability across languages, and assess their correlation with the \dat and other creativity tests to evaluate convergent and divergent validity. 

\begin{table*}[htbp]
  \centering
  \small
  \begin{tabular}{@{}llcl@{}}
      \toprule
      \textbf{Model Name} & \textbf{Provider} & \textbf{Multilingual} & \textbf{Reference} \\
      \midrule
      \textthreelarge\footnotemark[1] & OpenAI & \checkmark & \citet{roberta2019unsupervised} \\
      \nomic\footnotemark[2] & Nomic AI & \xmark & \citet{nussbaumNomicEmbed2024} \\
      \snowflake\footnotemark[3] & Snowflake & \xmark &  \citet{merrickArcticEmbed2024} \\
      \snowflaketwo\footnotemark[4] & Snowflake & \checkmark &  \citet{yuArcticEmbed2_2024} \\
      \granite\footnotemark[5] & IBM & \checkmark & \citet{awasthyGraniteEmbeddingModels2025} \\
      E5-Mistral-7B-instruct\footnotemark[6] & Microsoft & \checkmark & \citet{wangImprovingTextEmbeddings2024} \\
      multilingual-e5-large-instruct\footnotemark[7] & Microsoft & \checkmark & \citet{wangMultilingualE5Text2024} \\
      BGE-M3\footnotemark[8] & BAAI & \checkmark & \citet{chenBGE_M3_Embedding_2024} \\
      \bottomrule
  \end{tabular}
  \caption{\label{tab:embedding_models}Overview of tested embedding models, their providers, multilingual capabilities, and references.}
\end{table*}

\subsection{Semantic distances}

To quantify the semantic distance between two words, we first map each word to a high-dimensional vector using the chosen embedding model (see Table~\ref{tab:embedding_models} for an overview). Let $\vec{a}$ and $\vec{b}$ denote the embedding vectors for two words $a$ and $b$. The \emph{cosine similarity} between these vectors is defined as:
\begin{equation}
    \text{cosine\_similarity}(\vec{a}, \vec{b}) = \frac{\vec{a} \cdot \vec{b}}{\|\vec{a}\| \|\vec{b}\|}
\end{equation}
where $\vec{a} \cdot \vec{b}$ is the dot product and $\|\vec{a}\|$ is the Euclidean norm of vector $\vec{a}$. Cosine similarity ranges from $-1$ (opposite directions) to $1$ (identical directions), with $0$ indicating orthogonality (no similarity).

To obtain a measure of \emph{dissimilarity} (or: \emph{semantic distance}), we use:
\begin{equation}
    \text{dissimilarity}(\vec{a}, \vec{b}) = 1 - \text{cosine\_similarity}(\vec{a}, \vec{b})
\end{equation}
This transformation ensures that higher values correspond to greater semantic distance between the two words. Thus, for each pair of words in a response, we compute their embedding vectors, calculate the cosine similarity, and then derive the dissimilarity as $1$ minus the cosine similarity.

\subsection{Multilingual embeddings}
\label{sec:multilingual_embeddings}
As discussed above, we opted for using a multilingual transformer-based embedding model, which similarly to \glove provides embeddings of words in a high-dimensional vector space, i.e., a word is mapped to a vector of numbers. To identify a suitable model, we tested a range of (multilingual) embedding models, see \cref{tab:embedding_models}, to evaluate their embedding stability across languages and scripts. 

Our employed process for the evaluation of the embedding models was the following:
\begin{enumerate}
  \item We generated a list of $100$ nouns in English.
  \item We translated the list into the target languages (cf. Table~\ref{tab:languages}).
  \item We computed the pairwise dissimilarity scores between all (approximately $5000$) pairs of translated nouns using the embedding models; we removed the diagonal of the matrix to remove the self-similarity, as well as invalid embeddings.
  \item We computed the mean, median, and standard deviation of the dissimilarity scores.
\end{enumerate}

For comparison of the obtained dissimilarity scores to the original DAT scoring approach (as well as to each other), we calibrated the considered embeddings against the \dat's \glove model, by using the same $100$ nouns and matching mean and standard deviation of the dissimilarity scores via a linear transformation to those of the \dat. For our tests but also the later scoring, we transform all words into lower case for the languages that are case-sensitive, to avoid spurious effects from creative casing.

\begin{table*}[h!]
  \centering
  \small
  \begin{tabular}{@{}lllll@{}}
      \toprule
      \textbf{Language} & \textbf{Abbr.} & \textbf{Family} & \textbf{Branch} & \textbf{Script Used} \\
      \midrule
      English     & en & Indo-European & Germanic           & Latin                        \\
      Spanish     & es & Indo-European & Romance            & Latin                        \\
      French      & fr & Indo-European & Romance            & Latin                        \\
      German      & de & Indo-European & Germanic           & Latin                        \\
      Italian     & it & Indo-European & Romance            & Latin                        \\
      Dutch       & nl & Indo-European & Germanic           & Latin                        \\
      Portuguese  & pt & Indo-European & Romance            & Latin                        \\
      Polish      & pl & Indo-European & Slavic (West)      & Latin (with diacritics)      \\
      Russian     & ru & Indo-European & Slavic (East)      & Cyrillic                     \\
      Hindi       & hi & Indo-European & Indo-Aryan         & Devanagari                   \\
      Japanese    & ja & Japonic        & —                  & Kanji + Hiragana + Katakana  \\
      Arabic      & ar & Afro-Asiatic   & Semitic            & Arabic                       \\
      Czech       & cs & Indo-European  & Slavic (West)      & Latin (with diacritics)      \\
      Korean      & ko & Koreanic       & —                  & Hangul                       \\
      Chinese     & zh & Sino-Tibetan   & Sinitic            & Simplified/Traditional Han   \\
      \bottomrule
  \end{tabular}
  \caption{\label{tab:languages}Overview of languages used in the study, their ISO abbreviations, families, branches, and scripts.}
\end{table*}

\begingroup
\scriptsize
\footnotetext[1]{\url{https://platform.openai.com/docs/models/text-embedding-3-large}}
\footnotetext[2]{\url{https://huggingface.co/nomic-ai/nomic-embed-text-v1.5}}
\footnotetext[3]{\url{https://huggingface.co/Snowflake/snowflake-arctic-embed-l}}
\footnotetext[4]{\url{https://huggingface.co/Snowflake/snowflake-arctic-embed-l-v2.0}}
\footnotetext[5]{\url{https://huggingface.co/ibm-granite/granite-embedding-278m-multilingual}}
\footnotetext[6]{\url{https://huggingface.co/intfloat/e5-mistral-7b-instruct}}
\footnotetext[7]{\url{https://huggingface.co/intfloat/multilingual-e5-large-instruct}}
\footnotetext[8]{\url{https://huggingface.co/BAAI/bge-m3}}
\endgroup

\subsubsection{Tested multilingual embedding models.}

After extensive testing of state-of-the-art multilingual embedding models (see \cref{tab:embedding_models} for an overview and the MTEB leaderboard\footnote{\url{https://huggingface.co/spaces/mteb/leaderboard}} for benchmarks), we selected the \emph{\granite} model \citep{awasthyGraniteEmbeddingModels2025} as our base model for \sdat. It is an open-weight model, and it performed best in our tests. 

The {\granite} model is a large, transformer-based multilingual embedding model developed by IBM, which follows an XLM-RoBERTa \citep{roberta2019unsupervised} configuration. It features 12 layers and 12 attention heads, with an embedding size of 768 and an intermediate size of 3072, resulting in a total of 278 million parameters. The model supports a vocabulary of 250,002 tokens and can process sequences up to 512 tokens in length. Utilizing the GeLU activation function, this model is designed to generate high-quality, language-agnostic embeddings across a wide range of languages. Compared to its smaller variants, the 278M model offers increased representational capacity and depth, making it particularly well-suited for tasks requiring robust multilingual semantic understanding, such as cross-lingual creativity assessment in the \sdat framework.

The \granite model has been finetuned on a large multilingual corpus and supports a wide range of languages, including English (en), Arabic (ar), Czech (cs), German (de), Spanish (es), French (fr), Italian (it), Japanese (ja), Korean (ko), Dutch (nl), Portuguese (pt), and Chinese (zh), cf. Table~\ref{tab:languages}. While the \granite model is primarily targeting these 12 languages, it can also be fine-tuned on other languages that are covered by the XLM-RoBERTa vocabulary (see \citet{roberta2019unsupervised} for details), which includes approximately 100 languages in total. This allows for extending the \sdat to other scripts and languages (see \citet{awasthyGraniteEmbeddingModels2025} for details).

\begin{figure*}[htbp]
    \centering
    \begin{tabular}{cc}
        \includegraphics[width=0.48\textwidth]{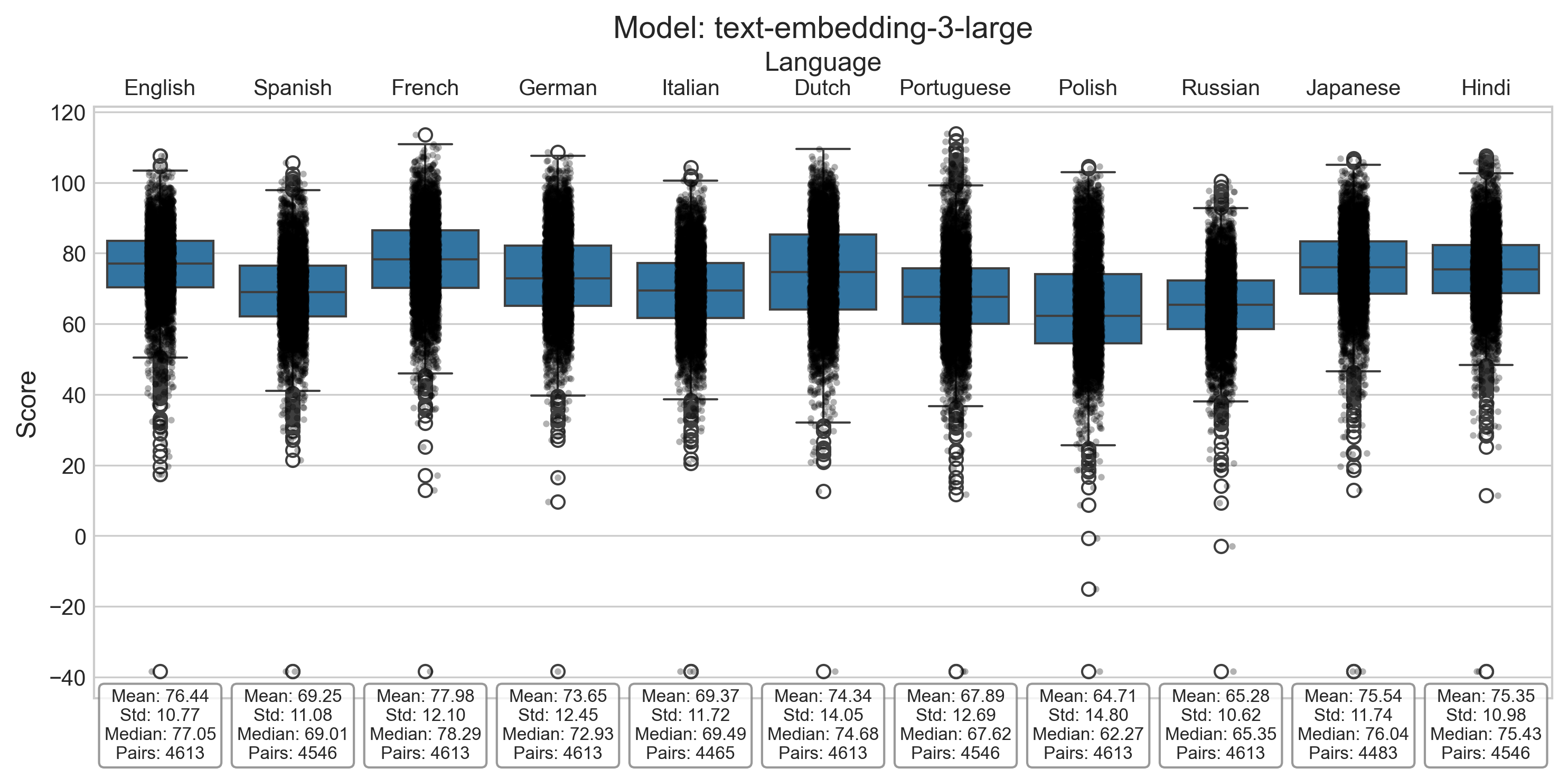} &
        \includegraphics[width=0.48\textwidth]{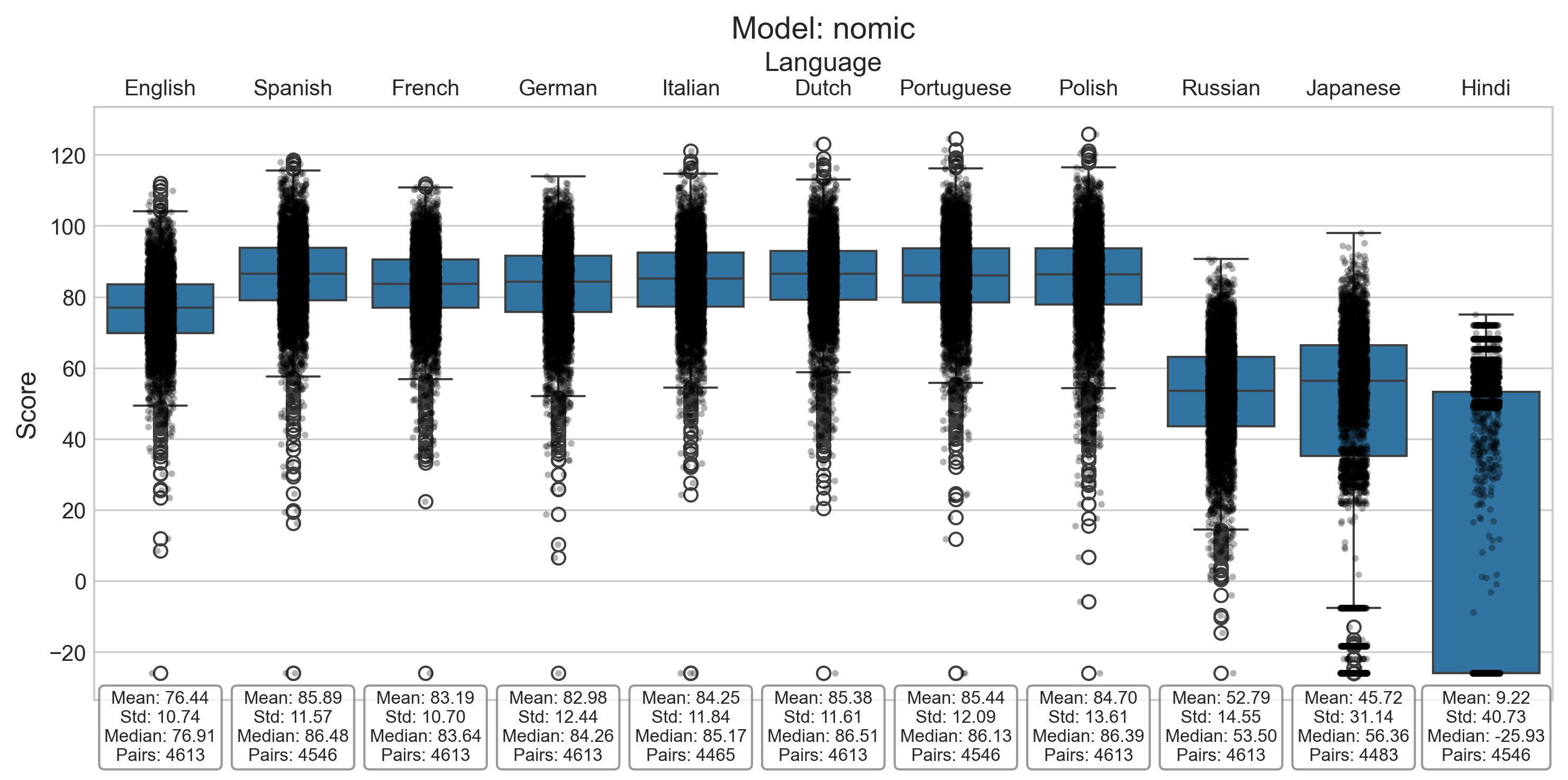} \\
        \includegraphics[width=0.48\textwidth]{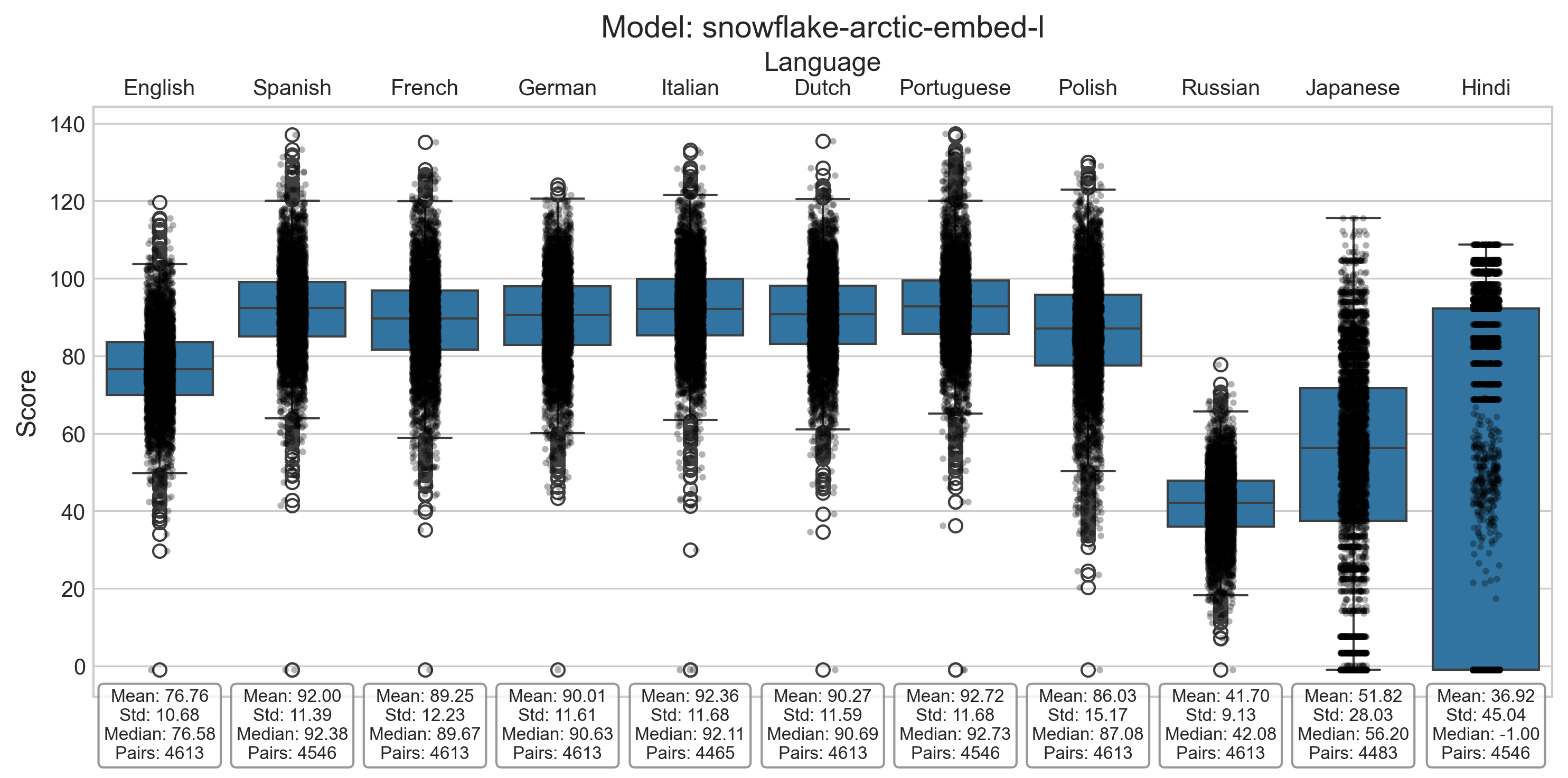} &
        \includegraphics[width=0.48\textwidth]{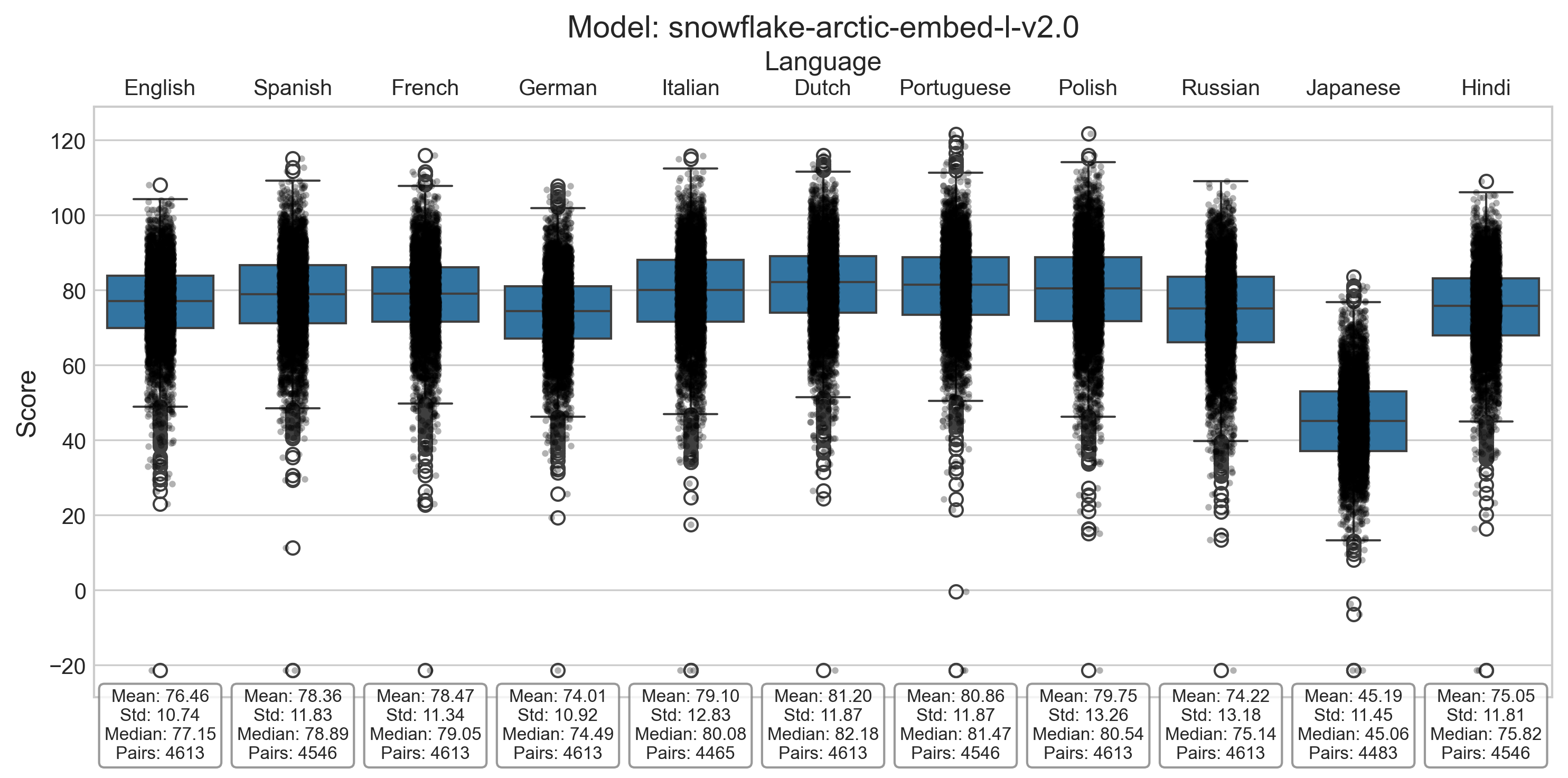} \\
        \includegraphics[width=0.48\textwidth]{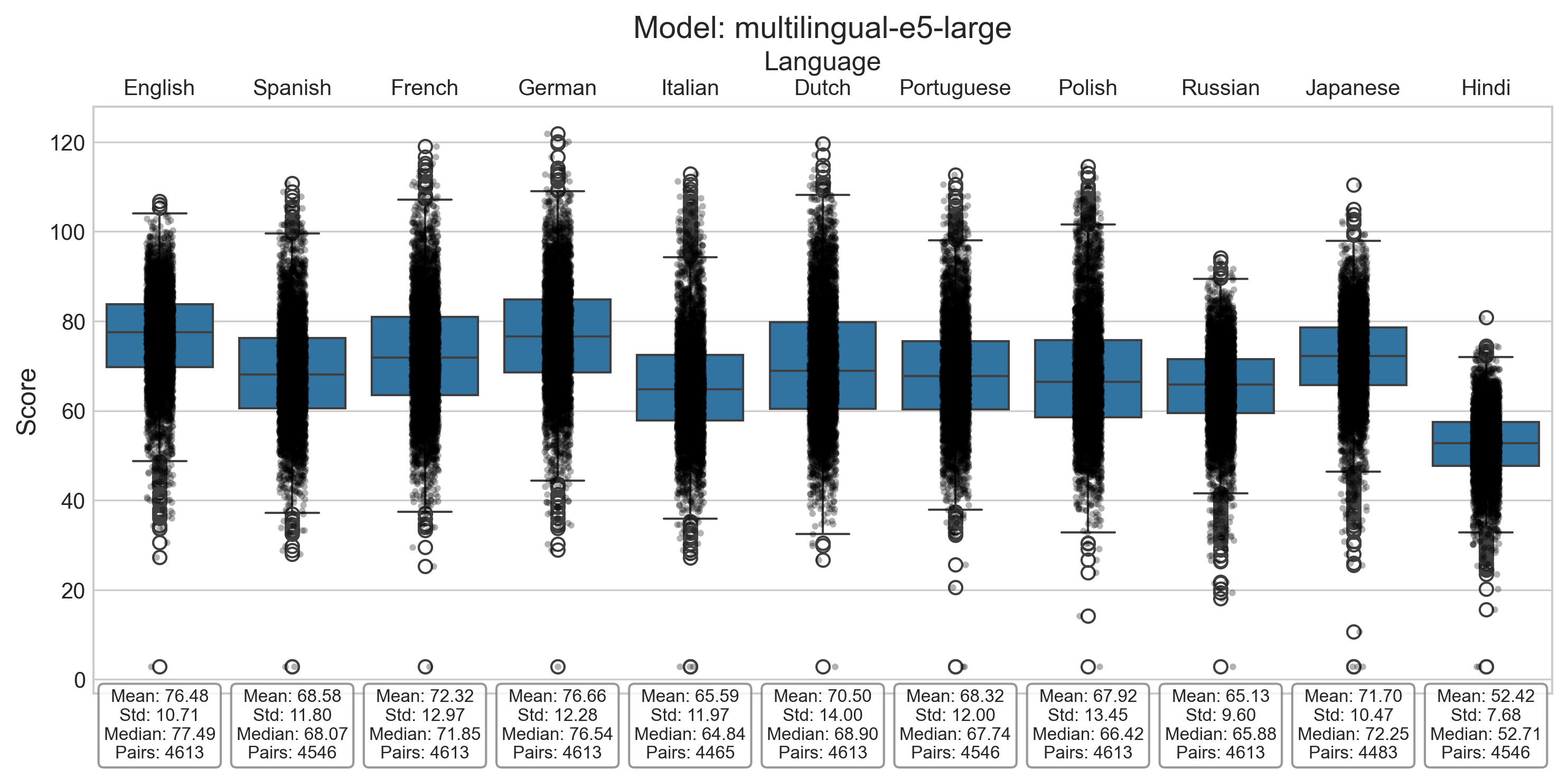} &
        \includegraphics[width=0.48\textwidth]{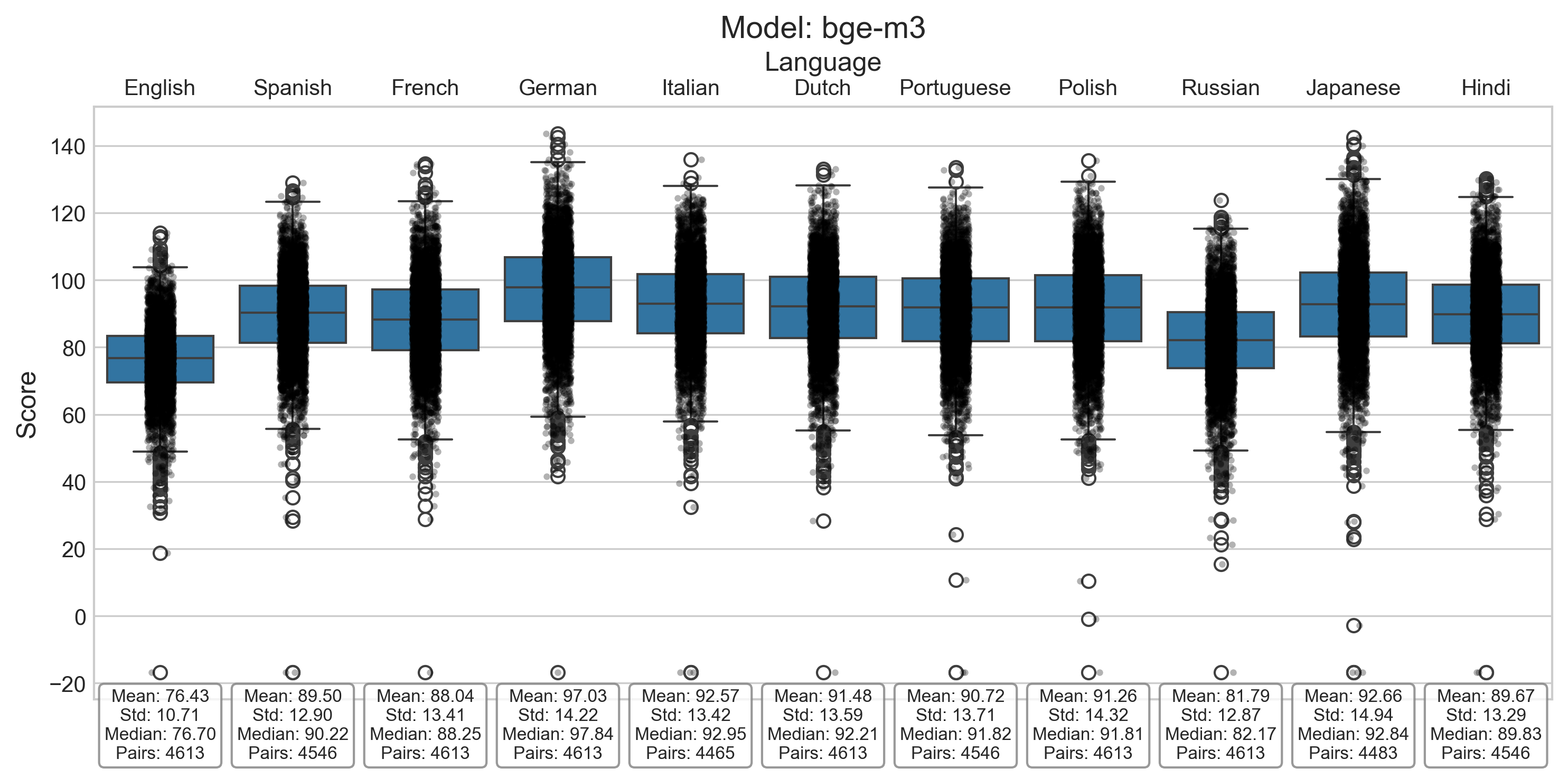} \\
        \includegraphics[width=0.48\textwidth]{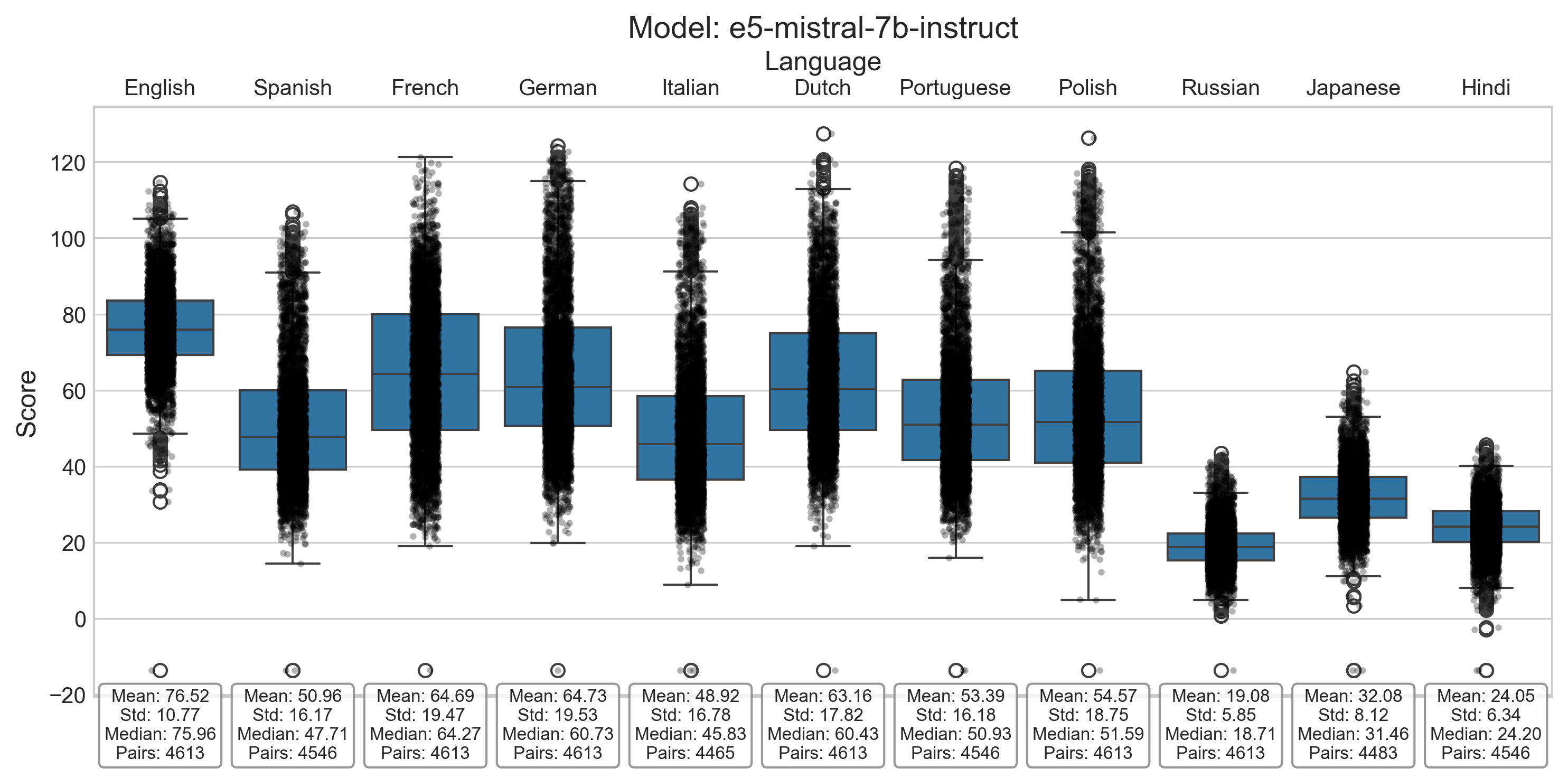} &
        \includegraphics[width=0.48\textwidth]{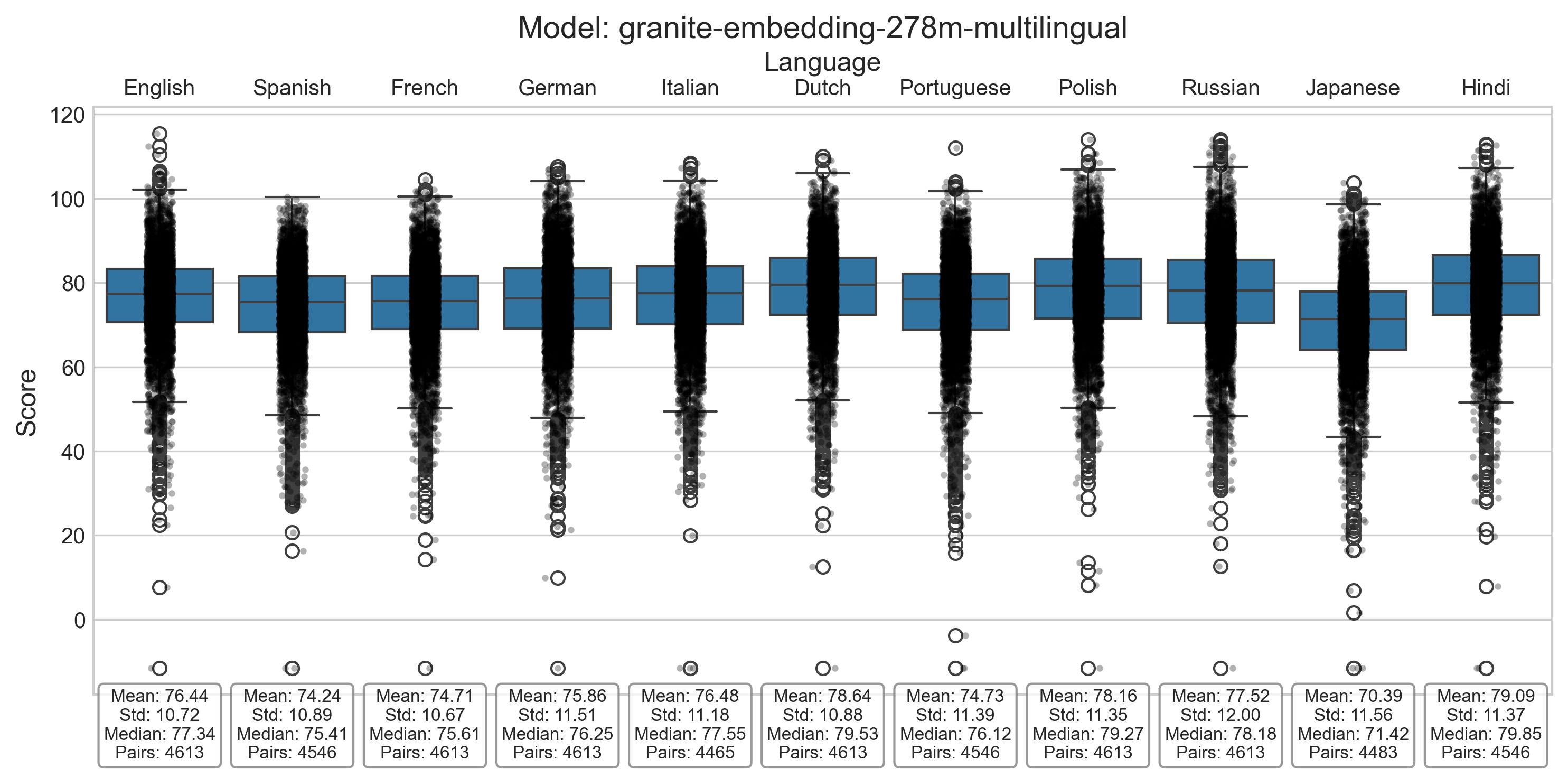} \\
    \end{tabular}
    \caption{Multilingual calibration results across different languages for the following models: \textthreelarge from OpenAI (row 1, left), \nomic (row 1, right), \snowflake (row 2, left), \snowflaketwo (row 2, right), multilingual-e5-large-instruct (row 3, left), BGE-M3 (row 3, right), E5-Mistral-7B-instruct (row 4, left), and \granite (row 4, right). While OpenAI's \textthreelarge model is a multilingual model, the pairwise dissimilarities of translated word pairs vary quite considerably across languages. The \nomic model is not a multilingual model, but rather trained for English language use, which leads to inflated pairwise dissimilarity values for non-English words written in the Latin script (used by languages such as English, German, Spanish, etc.) and for non-latin script (e.g., Russian, Chinese, Arabic, etc.) to significantly lower values and effectively collapses; the same holds for the \snowflake model. The \snowflaketwo model is a multilingual model and shows a more stable calibration across languages; however also shows significant degradation for Chinese character-based languages like Japanese and Chinese, shown here for Japanese. While multilingual-e5-large-instruct and BGE-M3 are multilingual models, their pairwise dissimilarities are rather unstable across languages. E5-Mistral-7B-instruct has no stable calibration across languages. \granite shows the best calibration across languages, including non-Latin script languages. Note that while \granite has not been trained for Hindi, it still generates a distribution similar to the languages it has been trained for, which is most likely due to using the same tokenization as \xlmroberta; whether this translates into proper semantic similarity scores will be investigated in future research. 
    }
    \label{fig:multilingual_calibration}
\end{figure*}

\begin{figure*}[htbp]
  \centering
  \begin{tabular}{ccc}
    \includegraphics[width=0.9\textwidth]{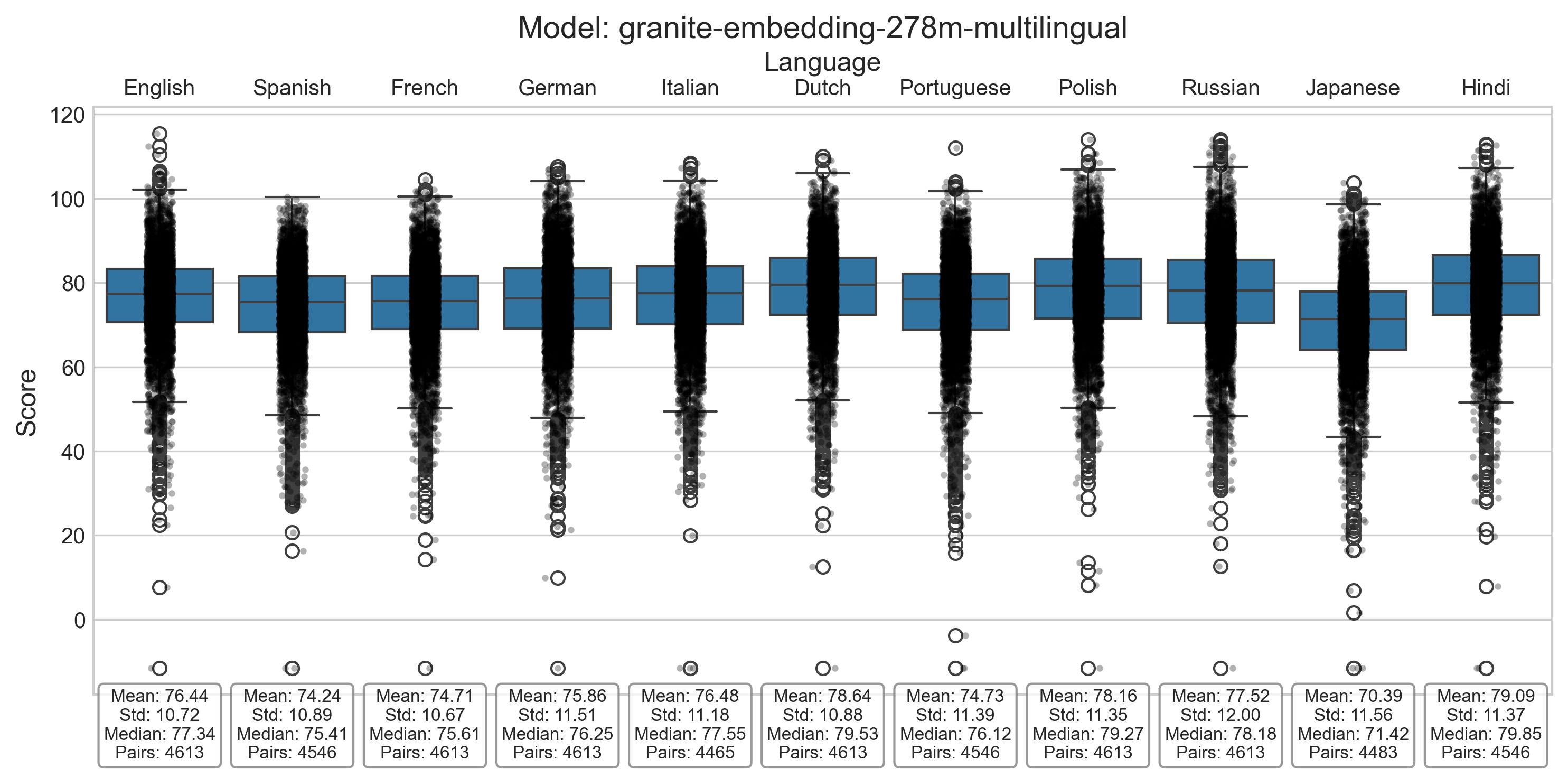} &
  \end{tabular}
  \caption{Multilingual calibration results across different languages for the \granite model. Here the distributions across languages (including non-Latin script languages) are rather stable and comparable, which is important for the \sdat.}
  \label{fig:multilingual_calibration_granite}
\end{figure*}

As seen in Fig.~\ref{fig:multilingual_calibration}and \ref{fig:multilingual_calibration_granite}, the \granite model's performance is very stable across different languages and scripts, even for those that it is not explicitly fine-tuned for. We believe that the rather robust semantic (dis-)similarity across languages that we observed in our tests arises from the inclusion of synthetically generated pairs in the training process (see \citet[Section 5]{{awasthyGraniteEmbeddingModels2025}}). However, we do observe a small shift in the distribution for Japanese, which we attribute to our way of translating the English nouns into kanji. Kanji however, often have multiple meanings with context (sentence and surrounding kanji) determining the particular one to use. This extra information is absent when testing via single words or simple compounds, as is done in the context of the \dat and \sdat.

\subsubsection{Percentile Calculation}
Raw scores alone provide limited insight into an individual's DT ability, as they lack context regarding the broader distribution of responses. Percentile scores, by contrast, position an individual's performance relative to the full distribution of test participants, offering a more interpretable measure of cognitive flexibility. To establish percentiles for the \sdat, we drew on multiple large-scale datasets from prior DAT studies. Specifically, we included data from the following sources:

\begin{itemize}
    \item \textbf{Study 1a, \citet{olson_naming_2021}}: 141 undergraduate psychology students from Melbourne, Australia, who completed the \dat as part of a cognitive assessment study. Participants also completed the AUT and the Bridge-the-Associative-Gap Task, a measure of convergent thinking.
    \item \textbf{Study 1b, \citet{olson_naming_2021}}: 285 undergraduate students from a similar demographic background, providing a more robust sample for score normalization. Participants also completed the AUT and the Bridge-the-Associative-Gap Task, a measure of convergent thinking.
    \item \textbf{Study 2, \citet{olson_naming_2021}}: A demographically diverse, large-scale sample (\textit{N} = 8,572) recruited via a national media campaign led by the Australian Broadcasting Corporation. This dataset includes a broad age range, with participants spanning from under 7 to over 70 years (data available at OSF\footnote{\url{https://osf.io/kbeq6/files/osfstorage}}). Participants also completed a very short version of the AUT.
\end{itemize}

For each dataset, raw \dat responses were re-scored using the \sdat framework. Percentile ranks were then calculated based on the aggregated distribution of these scores, allowing for meaningful comparisons across studies and participant demographics. This approach ensures that the resulting percentile scores are representative of a wide range of age groups, cultural backgrounds, and linguistic contexts, thereby supporting the cross-linguistic and cross-cultural ambitions of the \sdat.

\subsubsection{Correlation of the \dat scores}

\begin{figure*}[htbp]
    \centering
    \includegraphics[width=0.9\textwidth]{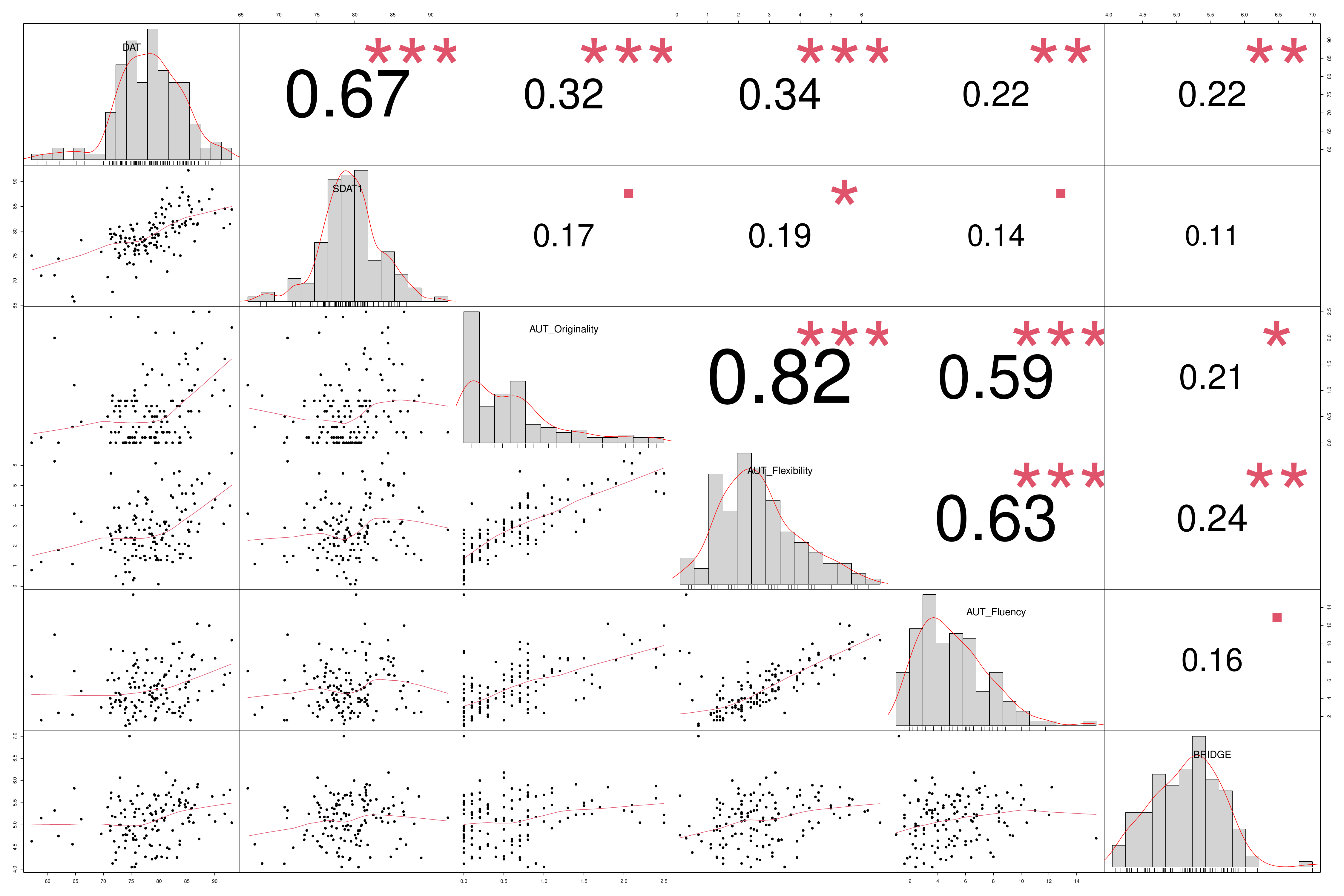}
    \caption{Correlation of embedding-based DAT scores with original \dat scores for the \sdat (based on \granite). The figure shows the relation between the calibrated \sdat scores and the original \dat scores, illustrating the effectiveness of the calibration process and the alignment of the new model with the established \dat score, as well as the Alternative Use Task (AUT) and the Bridge-the-Associative-Gap Task (Bridge).}
    \label{fig:dat_correlation}
\end{figure*}

As mentioned above, we calibrated the considered embeddings against the \dat's \glove model via a simple linear transformation to match the mean and standard deviation of the dissimilarity scores. In \cref{fig:dat_correlation} we depict the correlation of the \sdat.

\begin{table*}[!h]
\centering
\small
\begin{tabular}{llcc}
\toprule
\textbf{Study} & \textbf{Measure} & \textbf{DAT} & \textbf{\sdat} \\
\midrule
\multicolumn{4}{l}{\textbf{Olson et al. Study 1a (ns = 138--141)}} \\
 & DAT && .67*** [.56, .75]  \\
 & AUT: Originality & .32*** [.16, .46] & .17* [--.00, .17] \\
 & AUT: Flexibility & .34*** [.18, .48] & .19* [.03, .35] \\
 & AUT: Fluency     & .22**  [.06, .37] & .14* [--.03, .30] \\
 & Bridge-the-Associative-Gap Task & .22** [.06, .38] & .11 [--.06, .28] \\
\addlinespace
\multicolumn{4}{l}{\textbf{Olson et al. Study 1b (ns = 205--284)}} \\
 & DAT && .60*** [.52, .67]  \\
 & AUT: Originality & .32*** [.20, .43] & .24*** [.11, .36] \\
 & AUT: Flexibility & .35*** [.23, .46] & .27*** [.15, .39] \\
 & AUT: Fluency     & .30*** [.17, .41] & .16** [.03, .28] \\
 & Bridge-the-Associative-Gap Task & .23*** [.10, .36] & .08 [--.06, .22] \\
\addlinespace
\multicolumn{4}{l}{\textbf{Olson et al. Study 2 (ns = 355--8,498)}} \\
 & DAT && .65*** [.64, .66]  \\
 & AUT: Originality & .13** [.03, .23] & .13** [.02, .23] \\
\bottomrule
\end{tabular}
\caption{Comparisons between correlations of Olson et al's \dat-score with our newly developed \sdat}
\caption*{\textit{Note}. *p $<$ .05, **p $<$ .01, ***p $<$ .001 (all p-values one-tailed). 95\%-confidence intervals are shown in square brackets. We focus on the full dataset in Studies 1a and 1b. Olson et al. (2021) also reported correlations between the \dat and other creativity measures, which had been manually screened to identify participants who followed instructions more closely, but only in their Study 1a and 2. For Study 2, Olson et al. only provided data for “manually screened” AUT-originality scores for a small subset of their sample (n = 355).}
\label{tab:correlations}
\end{table*}

\begin{figure*}[!h]
    \centering
    \includegraphics[width=0.9\textwidth]{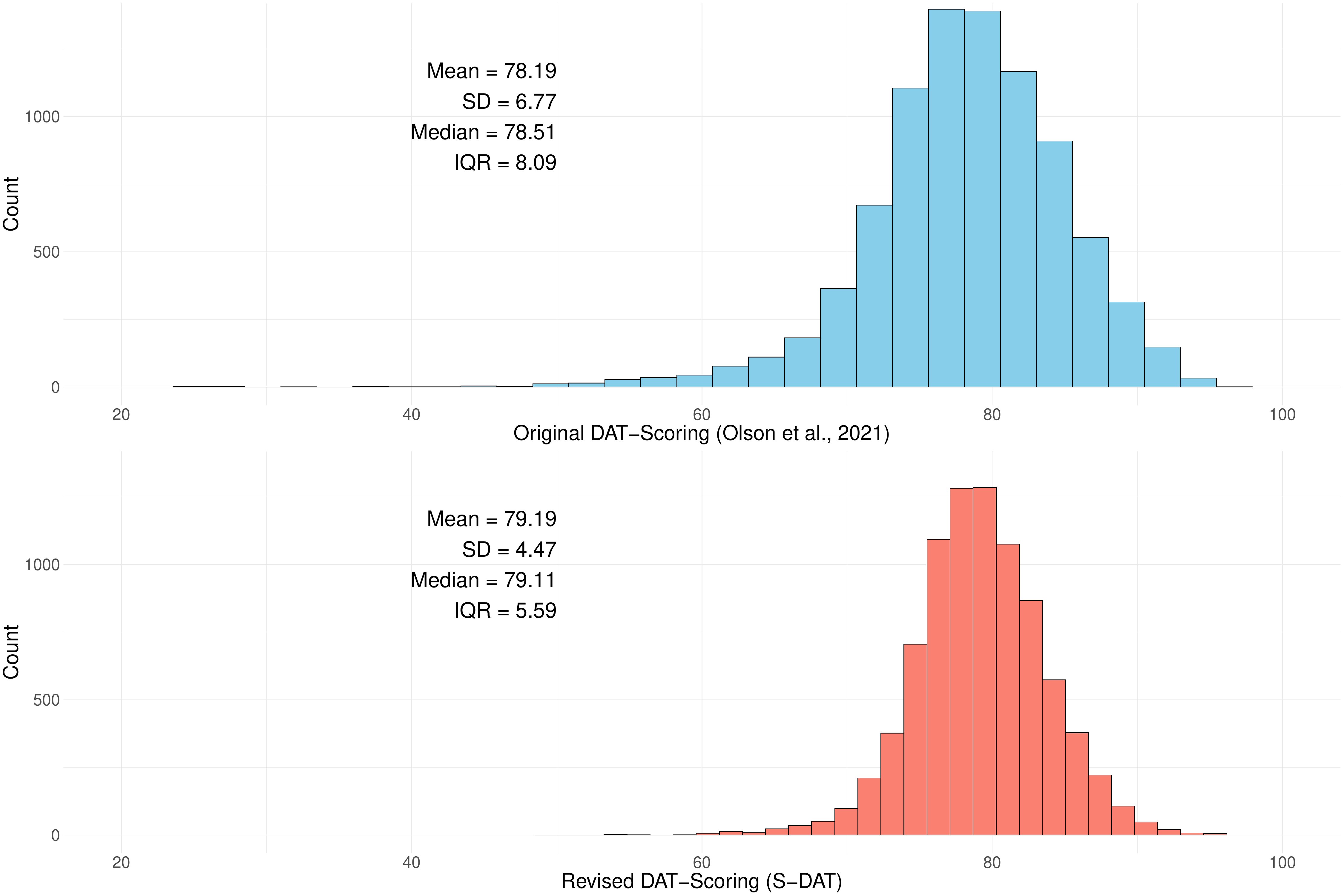}
    \caption{Histograms displaying the distribution of the original DAT-scores as well as the S-DAT when applied to the data from Olson et al. (2021), Study 2. While the \sdat's underlying \granite-based semantic distances have been calibrated to match the mean and standard deviation of the \dat's underlying \glove-based semantic distances, the \sdat's score distribution (i.e., scoring the 10 provided pairs) has slightly lower standard deviation and a slightly higher mean than the \dat's score distribution. In particular, the \sdat is slightly more robust to outliers; see also the transport graphic in Figure~\ref{fig:within}.
    }
    \label{fig:histograms}
\end{figure*}

\begin{figure*}[!h]
    \centering
    \includegraphics[width=0.9\textwidth]{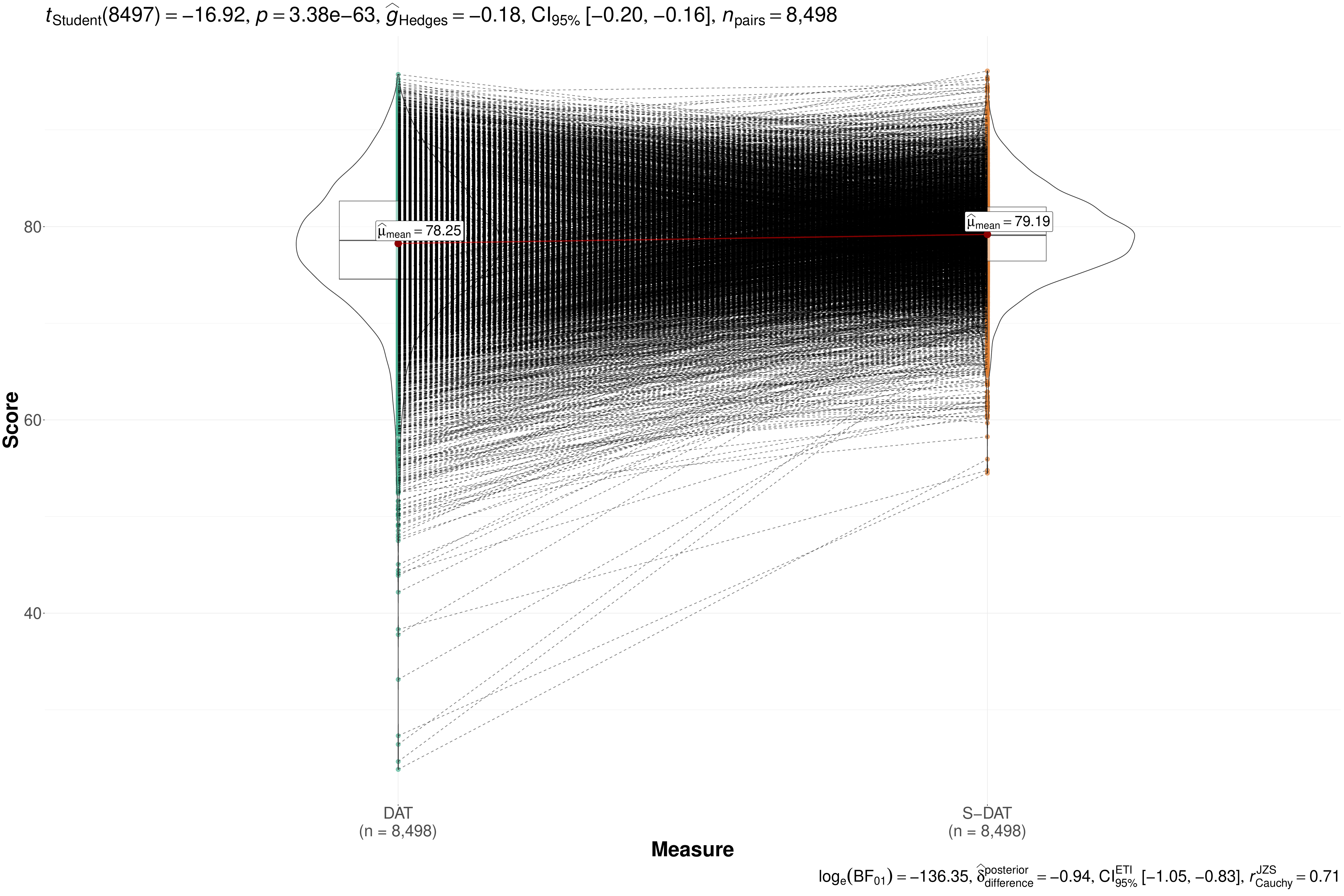}
    \caption{Comparison between the original DAT-scores as well as the S-DAT based on the data from \citet{olson_naming_2021}, Study 2.}
    \label{fig:within}
\end{figure*}

\subsubsection{Correlations with Human Creativity Measures.}
To validate the new score, we used the data provided by Olson et al. (2021, their Studies 1a, 1b, and 2, as shortly introduced above), which validated the original \dat using established creativity measures. In all studies, DT was assessed via the AUT: for Studies 1a and 1b, participants were presented with five common household objects (brick, paper clip, newspaper, ice tray, rubber band) and asked to list as many creative uses as possible within 2 minutes per item. Responses were scored on three dimensions: fluency (total number of uses), originality (rarity within the sample, scored from 0 to 3), and flexibility (number of distinct use categories). Interrater reliability for all measures was high (e.g., fluency \textit{r}$~>~.99$; flexibility \textit{r}~$=~.94–.97$). In Study 2, a shortened AUT was administered in which participants generated a single imaginative use for two objects, randomly drawn from a set. Two independent judges rated originality on a 1–5 scale (interrater reliability \textit{r}~$=~.66$).

To assess convergent thinking, participants in Studies 1a and 1b completed the Bridge-the-Associative-Gap Task. In this task, participants were shown two words (e.g., giraffe, scarf) and asked to generate a third word that semantically linked them (e.g., neck). Each trial was limited to 30 seconds, and participants responded to 20 word pairs (Study 1a) or the full 40-item set (Study 1b). Responses were scored for appropriateness on a 1–5 scale by two judges, with interrater reliability ranging from \textit{r}$~=~.67$ to $.78$.

\section{Results}
Overall, our newly developed DAT-Score, the \sdat, correlated similarly, albeit on average slightly less strongly, than the DAT-scores developed by \citet{olson_naming_2021} with the AUT-scores as well as the Bridge-the-Associative-Gap task (Tab. \ref{tab:correlations}). Interestingly, the correlations between the \sdat and the Bridge-the-Associative-Gap task, which measures convergent thinking, were non-significant, suggesting better discriminant validity of the \sdat compared to the \dat.
The overall distribution of scores was similar (Fig. \ref{fig:histograms} and Fig.~\ref{fig:within})), even though our \sdat score contained fewer outliers on both the higher but especially the lower end of the distribution. This can also be seen in the dispersion measures: The standard deviation and the inter-quartile range (IQR) were both smaller for the \sdat compared to the \dat (Fig. \ref{fig:histograms}). The 5th, 10th, 25th, 50th, 75th, 90th, and 95th percentiles also reflect the narrower distribution of the \sdat score compared to the \dat score (Tab.~\ref{tab:percentiles}). The data and R-code to reproduce these analyses can be found on OSF\footnote{\url{https://osf.io/pv84c/?view_only=7ba49b7b8cb74a2c92b91add66e7c72b}}.

\begin{table}[ht]
\centering
\small
\begin{tabular}{lrr}
\toprule
\textbf{Percentile} & \textbf{\dat } & \textbf{\sdat} \\
\midrule
5\%  & 66.99 & 72.17 \\
10\% & 70.51 & 73.98 \\
25\% & 74.52 & 76.44 \\
50\% & 78.51 & 79.11 \\
75\% & 82.61 & 82.03 \\
90\% & 86.28 & 84.87 \\
95\% & 88.45 & 86.59 \\
\bottomrule
\end{tabular}
\caption{Selected percentiles for \dat and \sdat based on $N$ = 8,572 participants from \citet{olson_naming_2021} Study 2.}
\label{tab:percentiles}
\end{table}

\section{Discussion}

The primary aim of this study was to develop a scalable, multilingual assessment of divergent thinking (DT) that overcomes the linguistic and logistical limitations of earlier approaches such as the original \dat \citep{olson_naming_2021}. The \sdat effectively addresses several key challenges in creativity assessment, including the need for cross-linguistic comparability, reduced participant burden, and scalable scoring, while also revealing important insights into the trade-offs involved in automated DT measurement.

One notable finding is that the \sdat shows slightly lower correlations with the AUT than the original \dat (see Table~\ref{tab:correlations}). While this might initially seem like a limitation, it actually aligns better with the construct validity of DT as a distinct cognitive process. The \sdat captures the associative breadth central to DT, while avoiding conflation with convergent thinking measures like the Bridge-the-Associative-Gap Task, which focus more on contextually appropriate solutions. This separation supports a more precise measurement of DT, reflecting its exploratory, open-ended nature \citep{guilford_creativity_1967, cropley_praise_2006}. The \sdat also demonstrates robust cross-linguistic performance for many languages, particularly those based on Latin scripts (e.g., English, Spanish, German, French), which share common linguistic structures. However, for languages with logographic or morphologically complex scripts, such as Japanese or Arabic, the straightforward word-pair approach of the \sdat may be less effective. For example, single characters in Japanese can have multiple meanings depending on context, which is not fully captured in static pairwise comparisons. Future studies should explore whether these differences necessitate language-specific norms to avoid cultural biases, as well as the potential for compound word strategies to distort percentile rankings in languages like German.

From an applied perspective, the \sdat offers several clear advantages over traditional creativity assessments. It is highly scalable, requires minimal participant effort, and can be administered quickly, making it ideal for large-scale studies and cross-cultural research. Unlike labor-intensive assessments such as the AUT or the assessment like the CAT, the \sdat reduces the time and resources needed for scoring, while maintaining consistency across linguistic contexts. This efficiency makes it particularly suitable for field studies, educational assessments, and experimental designs where rapid data collection is critical. However, it is important to recognize that the \sdat primarily captures novelty through semantic distance, without accounting for the contextual appropriateness or practical value of ideas. This reflects a broader challenge in automated creativity assessment: while semantic distance effectively measures conceptual remoteness, it does not fully capture the evaluative and pragmatic dimensions of creative thinking. As \citet{beaty_automating_2021} note, human creativity often involves not just the generation of novel associations, but also the refinement and contextualization of those ideas—a nuance that current embedding models struggle to capture.

Several limitations of the \sdat should be addressed in future work. First, while the current approach effectively captures semantic distance in many languages, it may require additional calibration for languages with more complex morphological structures or culturally specific semantic associations. Second, the reliance on static word-pair comparisons overlooks the dynamic, context-dependent nature of human associations, which are influenced by prior knowledge, emotional state, and situational context. Finally, the \sdat's focus on associative distance, while a valuable proxy for DT, may overemphasize novelty at the expense of other creativity dimensions like appropriateness, impact, or feasibility. Future research should explore hybrid approaches that integrate semantic distance with task-specific criteria or human ratings, potentially improving the construct validity of automated DT assessments.

In summary, the \sdat represents another step forward in automated, multilingual creativity assessment, providing a scalable, low-burden alternative to traditional methods. However, its effectiveness depends on the quality of the underlying embeddings, the cultural context of the target population, and the theoretical framing of creativity itself. Future work should focus on refining the linguistic and contextual sensitivity of the \sdat and expanding its applicability to non-Latin scripts.

\section{Acknowledgments}

We would like to thank the \href{https://www.zib.de}{Zuse Institute Berlin} for hosting various LLM models for testing and Peter Organisciak for providing us with an API key for the OpenScoring API. We would also like to thank the authors of \citet{olson_naming_2021} and \citet{organisciak_beyond_2023} for making their work and codes, including their scoring sites, publicly available. 

Research reported in this paper was partially supported by the Deutsche Forschungsgemeinschaft (DFG) through the DFG Cluster of Excellence MATH+ (grant number EXC-2046/1, project ID 390685689), and by the German Federal Ministry of Education and Research (BMBF), grant number 16DII133 (Weizenbaum-Institute).

\bibliographystyle{apalike} 
\bibliography{references,extras}

\end{document}